\documentclass{article}

\usepackage{times}
\usepackage{graphicx} 
\usepackage{subcaption} 

\usepackage{natbib}

\usepackage{algorithm}
\usepackage{algorithmic}

\usepackage{hyperref}

\usepackage{enumitem}

\usepackage[accepted]{icml2016_arxiv}

\usepackage{gensymb}
\usepackage{amsfonts}
\usepackage{multirow}
\usepackage{units}

\renewcommand{\abovespace}{\abovestrut{0.14in}}
\renewcommand{\belowspace}{\belowstrut{0.07in}}

\newcommand{\half}{\nicefrac{1}{2}}
\newcommand{\quarter}{\nicefrac{1}{4}}

\icmltitlerunning{Exploiting Cyclic Symmetry in CNNs}

\begin{document} 

\twocolumn[
\icmltitle{Exploiting Cyclic Symmetry in Convolutional Neural Networks}

\icmlauthor{Sander Dieleman}{sedielem@google.com}
\icmlauthor{Jeffrey De Fauw}{defauw@google.com}
\icmlauthor{Koray Kavukcuoglu}{korayk@google.com}

\icmladdress{Google DeepMind}

\icmlkeywords{symmetry, rotations, cyclic, dihedral, rotational symmetry, neural networks, convolutional neural networks}

\vskip 0.3in
]

\begin{abstract}
Many classes of images exhibit rotational symmetry. Convolutional neural networks are sometimes trained using data augmentation to exploit this, but they are still required to learn the rotation equivariance properties from the data. Encoding these properties into the network architecture, as we are already used to doing for translation equivariance by using convolutional layers, could result in a more efficient use of the parameter budget by relieving the model from learning them. We introduce four operations which can be inserted into neural network models as layers, and which can be combined to make these models partially equivariant to rotations. They also enable parameter sharing across different orientations. We evaluate the effect of these architectural modifications on three datasets which exhibit rotational symmetry and demonstrate improved performance with smaller models.
\end{abstract} 

\section{Introduction}
\label{sec:introduction}

For many machine learning applications involving sensory data (e.g. computer vision, speech recognition), neural networks have largely displaced traditional approaches based on handcrafted features. Such features require considerable engineering effort and a lot of prior knowledge to design. Neural networks however are able to automatically extract a lot of this knowledge from data, which previously had to be incorporated into models using feature engineering.

Nevertheless, this evolution has not exempted machine learning practitioners from getting to know their datasets and problem domains: prior knowledge is now encoded in the architecture of the neural networks instead. The most prominent example of this is the standard architecture of convolutional neural networks (CNNs), which typically consist of alternating convolutional layers and pooling layers. The convolutional layers endow these models with the property of translation \emph{equivariance}: patterns that manifest themselves at different spatial positions in the input will be encoded similarly in the feature representations extracted by these layers. The pooling layers provide local translation invariance, by combining the feature representations extracted by the convolutional layers in a way that is position-independent within a local region of space. Together, these layers allow CNNs to learn a hierarchy of feature detectors, where each successive layer sees a larger part of the input and is progressively more robust to local variations in the layer below.

CNNs have seen widespread use for computer vision tasks, where their hierarchical structure provides a strong prior that is especially effective for images. They owe their success largely to their scalability: the parameter sharing implied by the convolution operation allows for the capacity of the model to be used much more effectively than would be the case in a fully connected neural network with a similar number of parameters, and significantly reduces overfitting. It also enables them to be used for very large images without dramatically increasing the number of parameters.

\begin{figure}
\begin{center}
\includegraphics[width=1.0\columnwidth]{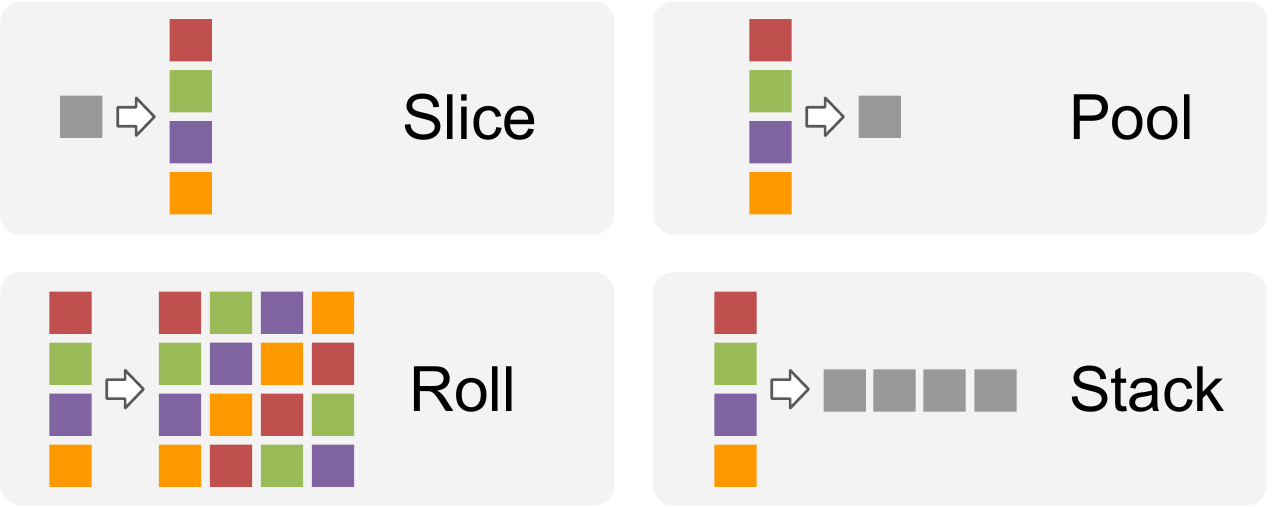}
\caption{The four operations that constitute our proposed framework for building rotation equivariant neural networks.}
\label{fig:framework}
\vspace{-10pt}
\end{center}
\end{figure} 

In this paper, we investigate how this idea can also be applied to rotation invariance and equivariance. Many types of data exhibit these properties, and exploiting them to increase parameter sharing may allow us to further regularise and scale up neural networks. We propose four new neural network layers, represented in Figure \ref{fig:framework}, which can be be used together to build CNNs that are partially or fully rotation equivariant. The resulting framework is scalable and easy to implement in practice.

\section{Cyclic symmetry}
\label{sec:cyclic-dihedral}

Filters in CNN layers learn to detect particular patterns at multiple spatial locations in the input. These patterns often occur in different orientations: for example, edges in images can be arbitrarily oriented. As a result, CNNs will often learn multiple copies of the same filter in different orientations.
This is especially apparent when the input data exhibits rotational symmetry. It may therefore be useful to encode a form of rotational symmetry in the architecture of a neural network, just like the parameter sharing resulting from the convolution operation encodes translational symmetry. This could reduce the redundancy of learning to detect the same patterns in different orientations, and free up model capacity. Alternatively it may allow us to reduce the number of model parameters and the risk of overfitting.

There are only four possible orientations of the input that allow for the application of a filter without interpolation: the rotations over angles  $k \cdot 90\degree, k \in \{0, 1, 2, 3\}$. This is because the sampling grid of an input rotated by one of these angles aligns with the original, which is not true for any other angle. We would like to avoid interpolation, because it adds complexity and can be a relatively expensive operation. If the input is represented as a matrix, the rotated versions can be obtained using only transposition and flipping of the rows or columns, two operations that are very cheap computationally. This group of four rotations is isomorphic to the cyclic group of order 4 ($C_4$), and we will refer to this restricted form of rotational symmetry as \emph{cyclic symmetry} henceforth\footnote{In literature, it is also known as `discrete rotational symmetry of order 4'.}.

In addition to four rotations, we can apply a horizontal flipping operation, for a total of eight possible orientations obtainable without interpolation. We will refer to this as \emph{dihedral symmetry}, after the dihedral group $D_4$.

\begin{figure}
\begin{center}
\includegraphics[width=1.05\columnwidth]{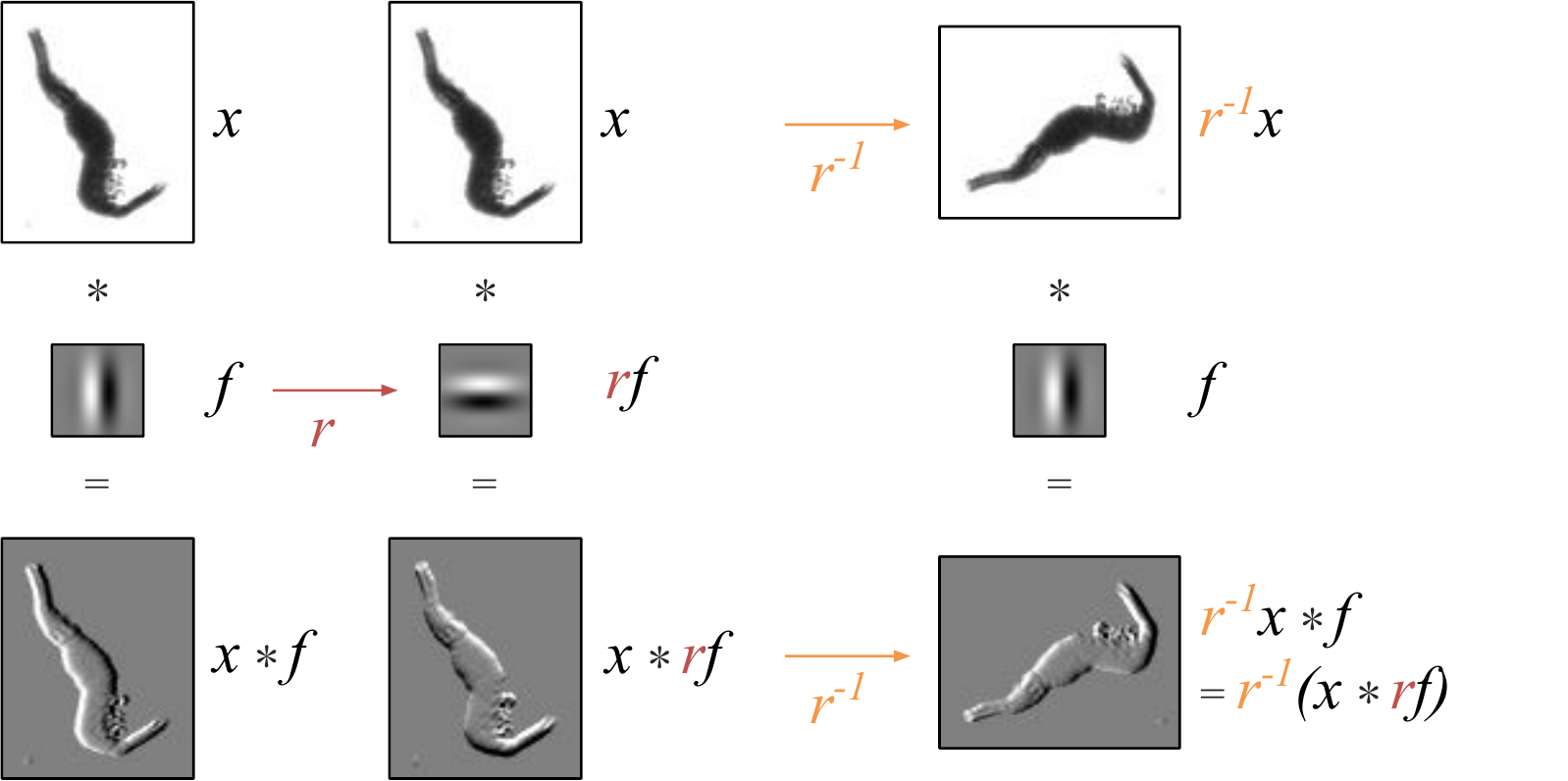}
\caption{Convolving an image with a rotated filter (middle) and inversely rotating the result is the same as convolving the inversely rotated image with the unrotated filter (bottom). This follows from the fact that rotation is distributive w.r.t. convolution.}
\label{fig:equivalence-filters-features}
\vspace{-10pt}
\end{center}
\end{figure}

We can encode cyclic symmetry in CNNs by parameter sharing: each filter should operate on four transformed copies of its input, resulting in four feature maps. Crucially, these feature maps are not rotated versions of each other, because the relative orientation of the input and the filter is different for each of them. This is equivalent to applying four transformed copies of the filter to the unchanged input. This is demonstrated in Figure \ref{fig:equivalence-filters-features}. In this paper we will primarily use the former interpretation. We discuss the practical implications of this choice in Section \ref{sec:rotate-filters-or-feature-maps}.

\section{Equivariance and invariance}
\label{sec:invariance-equivariance}

Many classes of images exhibit partial or full rotational symmetry, particularly in biology, astronomy, medicine and aerial photography. Some types of data specifically exhibit dihedral symmetry, such as board configurations in the game of Go. The tasks which we wish to perform on such data with neural networks usually require \emph{equivariance} to rotations: when the input is rotated, the learnt representations should change in a predictable way \cite{lenc2014understanding}. More formally, a function $f$ is equivariant to a class of transformations $\mathcal{T}$, if for all transformations $\mathbf{T} \in \mathcal{T}$ of the input $\mathrm{x}$, a corresponding transformation $\mathbf{T}'$ of the output $f(\mathrm{x})$ can be found, so that $f(\mathbf{T}\mathrm{x}) = \mathbf{T}'f(\mathrm{x})$ for all $\mathrm{x}$ \cite{schmidt2012learning}.

Often, the representations should not change at all when the input is rotated, i.e. they should be \emph{invariant}. It follows that an invariant representation is also equivariant (but not necessarily vice versa). In this case, $\mathbf{T}'$ is the identity for all $\mathbf{T}$. We also consider the special case where $\mathbf{T}' = \mathbf{T}$, i.e. the transformations of the input and the output are the same. We will refer to this as \emph{same-equivariance}.

\section{Encoding equivariance in neural nets}
\label{sec:encoding-equivariance}

The simplest way to achieve (approximate) invariance to a class of transformations of the input, is to train a neural network with \emph{data augmentation} \cite{simard2003best}:
during training, examples are randomly perturbed with transformations from this class, to encourage the network to produce the correct result regardless of how the input is transformed. Provided that the network has enough capacity, it should be able to learn such invariances from data in many cases \cite{lenc2014understanding}. But even if it perfectly learns the invariance on the training set, there is no guarantee that this will generalise. To obtain such a guarantee, we might want to encode the desired invariance properties in the network architecture, and allow it to use the additional freed up learning capacity to learn other concepts. Forcing the network to rediscover prior knowledge that we have about the data is rather wasteful.

To obtain more invariant predictions from a CNN, a straightforward approach is to produce predictions for a number of different transformations of the input, and to simply combine them together by averaging them. This requires no modifications to the network architecture or the training procedure, but comes at the expense of requiring more computation to generate predictions (one inference step for each transformation considered). This approach can also be applied in the case of partial rotation invariance, or as part of a broader \emph{test-time augmentation} strategy, which includes other transformations besides rotation over which predictions are averaged. However, as we discussed before in Section \ref{sec:encoding-equivariance}, it does not provide any guarantees about the invariance properties of the resulting model and it does not solve the problem in a principled way.

For the remainder of this section we will discuss the case of cyclic symmetry only (i.e. the class of transformations consisting of the rotations over $k \cdot 90\degree, k \in \{0, 1, 2, 3\}$), but the proposed framework can be generalised to other cases (including but not limited to dihedral symmetry).

\begin{figure*}
\begin{center}
\includegraphics[width=1.0\textwidth]{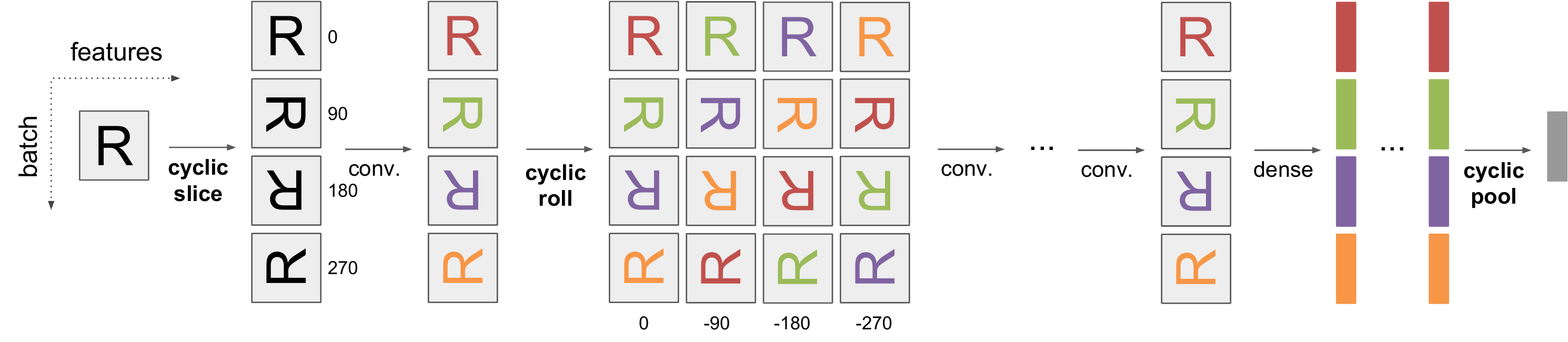}
\caption{Schematic representation of the effect of the cyclic slice, roll and pool operations on the feature maps in a CNN. Arrows represent network layers. Each square represents a minibatch of feature maps. The letter `R' is used to clearly distinguish orientations. Different colours are used to indicate that feature maps are qualitatively different, i.e. they are not rotations of each other. Feature maps in a column are stacked along the batch dimension in practice; feature maps in a row are stacked along the feature dimension.}
\label{fig:slice-roll-pool}
\vspace{-10pt}
\end{center}
\end{figure*}

\subsection{Framework}

\begin{table*}
\caption{The four operations that form our framework for building neural networks that are equivariant to cyclic rotations. Let $\mathbf{x} = [\mathrm{x}_0, \mathrm{x}_1, \mathrm{x}_2, \mathrm{x}_3]^T$, $r$ be the clockwise rotation over $90\degree$, $\sigma$ be a cyclic permutation shifting the elements backward, and $p$ be any permutation-invariant pooling function.}
\label{tab:framework}
\begin{center}
\begin{small}
\begin{tabular}{cp{6cm}cc}
\hline
\abovespace\belowspace
\sc{Name} & \sc{Definition} & \sc{Batch size} & \sc{\# feature maps} \\ 
\hline
\abovespace
\belowspace
Slice     & $S(\mathrm{x}) = [\mathrm{x}, r\mathrm{x}, r^2\mathrm{x}, r^3\mathrm{x}]^T$                               & $\times 4$ & unchanged \\
\belowspace
Pool      & $P(\mathbf{x}) = p(\mathrm{x}_0, r^{-1}\mathrm{x}_1, r^{-2}\mathrm{x}_2, r^{-3}\mathrm{x}_3)$             & $\div 4$   & unchanged \\
\belowspace
Stack     & $T(\mathbf{x}) = [\mathrm{x}_0, r^{-1}\mathrm{x}_1, r^{-2}\mathrm{x}_2, r^{-3}\mathrm{x}_3]$              & $\div 4$   & $\times 4$ \\
\belowspace
Roll      & $R(\mathbf{x}) = [T(\mathbf{x}), T(\sigma \mathbf{x}), T(\sigma^2 \mathbf{x}), T(\sigma^3 \mathbf{x})]^T$ & unchanged  & $\times 4$ \\
\hline
\end{tabular}
\end{small}
\end{center}
\vskip -0.1in
\end{table*}

We will introduce four operations, which can be cast as layers in a neural network, that constitute a framework that we can use to easily build networks that are equivariant to cyclic rotations and share parameters across different orientations. An overview is provided in Table \ref{tab:framework} and a visualisation in Figure \ref{fig:framework}. Each of the operations changes the size of the minibatch (slicing, pooling), the number of feature maps (rolling), or both (stacking). Beyond this, these operations do not affect the behaviour of the surrounding layers in any way, so in principle they are compatible with recent architectural innovations such as Inception \cite{szegedy2014going} and residual learning \cite{he2015deep}.

\subsection{Cyclic slicing and pooling}
To encode rotation equivariance in a neural network architecture, we introduce two new types of layers:
\begin{itemize}[leftmargin=1em]
 \item the \textbf{cyclic slicing} layer, which stacks rotated copies of a set of input examples into a single minibatch (which is $4\times$ larger as a result);
 \item the \textbf{cyclic pooling} layer, which combines predictions from the different rotated copies of an example using a permutation-invariant pooling function (e.g. averaging), reducing the size of the minibatch by $4\times$ in the process.
\end{itemize}

Formally, let $\mathrm{x}$ be a tensor representing a minibatch of input examples or feature maps, and let $r$ be the clockwise rotation over $90\degree$. The cyclic slicing operation can then be represented as $S(\mathrm{x}) = [\mathrm{x}, r\mathrm{x}, r^2\mathrm{x}, r^3\mathrm{x}]^T$, where we use a column vector to indicate that the rotated feature maps are stacked across the batch dimension in practice. As a result, all layers following the slicing layer will process each data point along four different \emph{pathways}, each corresponding to a different relative orientation of the input.

Let $\mathbf{x} = [\mathrm{x}_0, \mathrm{x}_1, \mathrm{x}_2, \mathrm{x}_3]^T$, then we define the pooling operation $P(\mathbf{x}) = p(\mathrm{x}_0, r^{-1}\mathrm{x}_1, r^{-2}\mathrm{x}_2, r^{-3}\mathrm{x}_3)$, where $p$ is a permutation-invariant pooling function, such as the average or the maximum. In practice, we find that some functions work better than others, as discussed in Section \ref{sec:pooling-functions}.

The pooling operation can be applied after one or more dense layers, at which point the feature maps no longer have any spatial structure. In that case, the inverse rotations that realign the feature maps can be omitted. The resulting network will then be \emph{invariant} to cyclic rotations. For some problems (e.g. segmentation) however, the spatial structure should be preserved up until the output, in which case the realignment is important. The resulting network will then be \emph{same-equivariant}: if the input rotates, the output will rotate in the same way.

Using these two layers, it is straightforward to modify an existing network architecture to be invariant or same-equivariant by inserting a slicing layer at the input side and a pooling layer at the output. It is important to note that the effective batch size for the layers in between will become $4\times$ larger, so we may also have to reduce the input batch size to compensate for this. Otherwise this modification would significantly slow down training.

We need not necessarily insert the pooling layer at the very end: we could also position some other network layers after the pooling operation, but not convolutional or spatial pooling layers. Otherwise, we would relinquish equivariance at that point: the pooled feature maps would rotate with the input (same-equivariance), so their relative orientation w.r.t. the filters of the following layers would change.

\subsection{Cyclic rolling}
Next, we introduce the \textbf{cyclic rolling} operation. We observe that each minibatch of intermediate activations in a network with cyclic slicing and pooling contains four sets of feature maps for each example. These are not just rotations of each other, as they correspond to different relative orientations of the filters and inputs (indicated by different colours in Figure \ref{fig:slice-roll-pool}). By \emph{realigning} and stacking them along the feature dimension, we can increase the number of feature maps within each pathway fourfold with a simple copy operation, which means the next convolutional layer receives a richer representation as its input. Because of this, we can reduce the number of filters in the convolutional layers while still retaining a rich enough representation. In practice, this amounts to 4-way parameter sharing: each filter produces not one, but four feature maps, resulting from different relative orientations w.r.t. the input.

To formalise this operation, we will first characterise the equivariance properties of the slicing operation $S$. When applied to a rotated input $r\mathrm{x}$, we obtain $S(r\mathrm{x}) = [r\mathrm{x}, r^2\mathrm{x}, r^3\mathrm{x}, \mathrm{x}]^T = \sigma S(\mathrm{x})$, where $\sigma$ denotes a cyclic permutation, shifting the elements backward along the batch dimension.

We define the stacking operation $T(\mathbf{x}) = [\mathrm{x}_0, r^{-1}\mathrm{x}_1, r^{-2}\mathrm{x}_2, r^{-3}\mathrm{x}_3]$, which realigns the feature maps $\mathrm{x}_i$ corresponding to the different pathways and stacks them along the feature dimension, resulting in a row vector\footnote{Note the similarity to the pooling operation, but now the realigned feature maps are stacked rather than combined together.}.

The roll operation $R$ then simply consists of applying $T$ to all possible cyclic permutations of the input, and stacking the results along the batch dimension: $R(\mathbf{x}) = [T(\mathbf{x}), T(\sigma \mathbf{x}), T(\sigma^2 \mathbf{x}), T(\sigma^3 \mathbf{x})]^T$, or equivalently, $R(\mathbf{x}) = [\mathbf{x}, \sigma r^{-1} \mathbf{x}, \sigma^2 r^{-2} \mathbf{x}, \sigma^3 r^{-3} \mathbf{x}]$.
Figure \ref{fig:slice-roll-pool} shows the effect of the cyclic slice, roll and pool operations on the feature maps in a CNN.

By stacking the feature maps in the right order, we are able to preserve equivariance across layers: $R$ is same-equivariant w.r.t. the cyclic permutation $\sigma$ ($R(\sigma \mathbf{x}) = \sigma R(\mathbf{x})$).
The resulting increase in parameter sharing can be used either to significantly reduce the number of parameters (and hence the risk of overfitting), or to better use the capacity of the model if the number of parameters is kept at the same level. In a network where a rolling layer is introduced after every convolutional layer, we can keep the number of parameters approximately constant by reducing the number of filters by half. This will in turn increase the number of produced feature maps by a factor of two, which ends up balancing out the number of parameters for each layer because it is proportional to both the number of input feature maps and the number of filters.

\subsection{Cyclic stacking}
We may also want to achieve parameter sharing by rolling in networks that are not required to be fully equivariant: as mentioned in Section \ref{sec:introduction}, even networks trained on natural images often exhibit redundancy in the filters learned in the first couple of layers. To accommodate this use case, we can simply \textbf{stack} (i.e. concatenate) feature maps obtained from the different orientations along the feature dimension at some point in the network, instead of pooling them together as we would otherwise. This corresponds to the stacking operation $T$ that we introduced previously.

\subsection{Rotate feature maps or filters?}
\label{sec:rotate-filters-or-feature-maps}

As mentioned in Section \ref{sec:cyclic-dihedral}, we can equivalently rotate either the filters or feature maps on which they operate to achieve 4-way parameter sharing, because only their relative orientation affects the result of the convolution. This implies that there are two possible practical implementations, both with their own advantages and disadvantages.

Rotating the feature maps may not seem to be the most natural choice, but it is the easiest to implement in many modern deep learning software frameworks because the roll operation can be isolated and viewed as a separate \emph{layer} in a neural network model. By stacking the feature maps corresponding to different orientations in the batch dimension, it becomes easier to exploit data parallelism (the effective batch size for most of the computationally expensive operations is larger). The feature maps must necessarily be square, otherwise it would not be possible to stack the different orientations together in a single batch. The filters need not be, however. It also implies that the roll operation has to make four copies of each feature map, which increases memory requirements.

Rotating the filters, on the other hand, means that the feature maps need not be square, but the filters must be. This operation cannot be isolated from the convolutional layers in a model, because it affects its parameters rather than the input activations. In many frameworks this complicates the implementation, and may require partial reimplementation of the convolutional layer abstraction. It only requires copying the filters, which are generally smaller than the feature maps on which they operate, so memory requirements are reduced. After training, it is straightforward to produce a version of the model that does not require any special operations beyond slicing and pooling, by stacking the different orientations of the filters for each convolutional layer. This version can then be used to perform inference on (non-square) inputs of any size.

Our choice for the former primarily stems from practical considerations (i.e. ease of implementation).

\subsection{Dihedral symmetry}
\label{sec:dihedral-symmetry}

The previous discussion readily generalises to the dihedral case, by changing the slice operation to include flipped in addition to rotated copies of the input (for a total of 8 copies), and by adapting all other operations accordingly. One complication is that the equivariance properties of dihedral slicing are less straightforward: the resulting permutation is no longer cyclic. It is also important to take into account that flipping and rotation do not commute.

\section{Related work}
While convolutional structure has become the accepted approach of encoding translational equivariance in image representations, there is no such consensus for other classes of transformations. Many architectural modifications have been proposed to encode rotation equivariance. \citet{schmidt2012learning} modify Markov random fields (MRF) to learn rotation-invariant image priors. \citet{kivinen2011transformation}, \citet{sohn2012learning} and \citet{schmidt2012learning} propose modified restricted Boltzmann machines (RBMs) with tied weights to achieve rotation equivariance.

\citet{fasel2006rotation} create multiple rotated versions of images and feed them to a CNN with filters shared across different orientations. The representations are gradually pooled together across different layers, yielding a fully invariant representation at the output. Their approach of rotating the input rather than the filters is identical to ours, but our strategy for achieving invariance using a single pooling layer allows the intermediate layers to use more accurate orientation information. \citet{dieleman2015rotation} also create multiple rotated and flipped versions of images and feed them to the same stack of convolutional layers. The resulting representations are then concatenated and fed to a stack of dense layers. While this does not yield invariant predictions, it does enable parameter sharing across orientations. In our framework, this approach can be reproduced using a slicing layer at the input, and a stacking layer between the convolutional and dense parts of the network. A similar approach is investigated by \citet{teney2016learning}, where filters of individual convolutional layers are constrained to be rotations of each other.

\citet{wu2015flip} apply rotated and flipped copies of each filter in a convolutional layer and then max-pool across the resulting activations. We concatenate them instead and prefer to pool only at the output side of the network to be able to achieve global equivariance, which is not possible if there are multiple pooling stages in the network. Indeed, they find that it is only useful to apply their approach in the higher layers of the network, and only to a subset of the filters so that some orientation information is preserved. \citet{clark2015training} force the filters of convolutional layers to exhibit dihedral symmetry through weight sharing, which means the resulting feature maps will necessarily be invariant. However, the network is only able to accurately detect fully symmetric patterns in the input, which is too restrictive for many problems. \citet{sifre2013rotation} propose a model resembling a CNN with fixed rather than learned filters, which is scaling- and rotation-invariant in addition to translation-invariant. \citet{gens2014deep} propose deep symmetry networks, a generalisation of CNNs that can form feature maps over arbitrary transformation groups.

We can also modify the architecture to facilitate learning of equivariance properties from data, rather than directly encode them. This approach is more flexible, but it requires more training data. The model of \citet{kavukcuoglu2009learning} is able to learn local invariance to arbitrary transformations by grouping filters into overlapping neighbourhoods whose activations are pooled together. \citet{liao2013learning} describe a template-based approach that successfully learns representations invariant to both affine and non-affine transformations (e.g. out-of-plane rotation). \citet{cohen2014learning} propose a probabilistic framework to model the transformation group to which a given dataset exhibits equivariance.
Tiled CNNs \cite{ngiam2010tiled}, in which weight sharing is reduced, are able to approximate more complex local invariances than regular CNNs. 

A third alternative to encoding or learning equivariance properties involves explicitly estimating the transformation applied to the input separately for each example, as in transforming auto-encoders \cite{hinton2011transforming} and spatial transformer networks \cite{jaderberg2015spatial}. This approach was also investigated for face detection by \citet{rowley1998rotation}.

Both \citet{goodfellow2009measuring} and \citet{lenc2014understanding} discuss how the equivariance properties of representations can be measured. The latter also show that representations learnt in the lower layers of CNNs are approximately linearly transformed when the input is rotated or flipped, which implies equivariance.

Concurrently with our work, \citet{cohen2016group} present group-equivariant convolutional neural networks. They provide a theoretically grounded formalism for exploiting symmetries in CNNs which describes the same type of models that can also be built with our framework.

\section{Experiments}

\subsection{Datasets}

\paragraph{Plankton} The Plankton dataset \cite{Cowen2015} consists of 30,336 grayscale images of varying size, divided unevenly into 121 classes which correspond to different species of plankton. We rescaled them to 95$\times$95 based on the length of their longest side. We split this set into separate validation and training sets of 3,037 and 27,299 images respectively. This dataset was used for the National Data Science Bowl\footnote{https://www.kaggle.com/c/datasciencebowl}, a data science competition hosted on the Kaggle platform. Although 130,400 images were provided for testing, their labels were never made public. These images were used to evaluate the competition results. We were able to obtain test scores by submitting predictions to Kaggle, even though the competition had already ended.

\begin{figure}
        \centering
        \begin{subfigure}[b]{0.16\textwidth}
		\centering
                \includegraphics[width=0.75\textwidth]{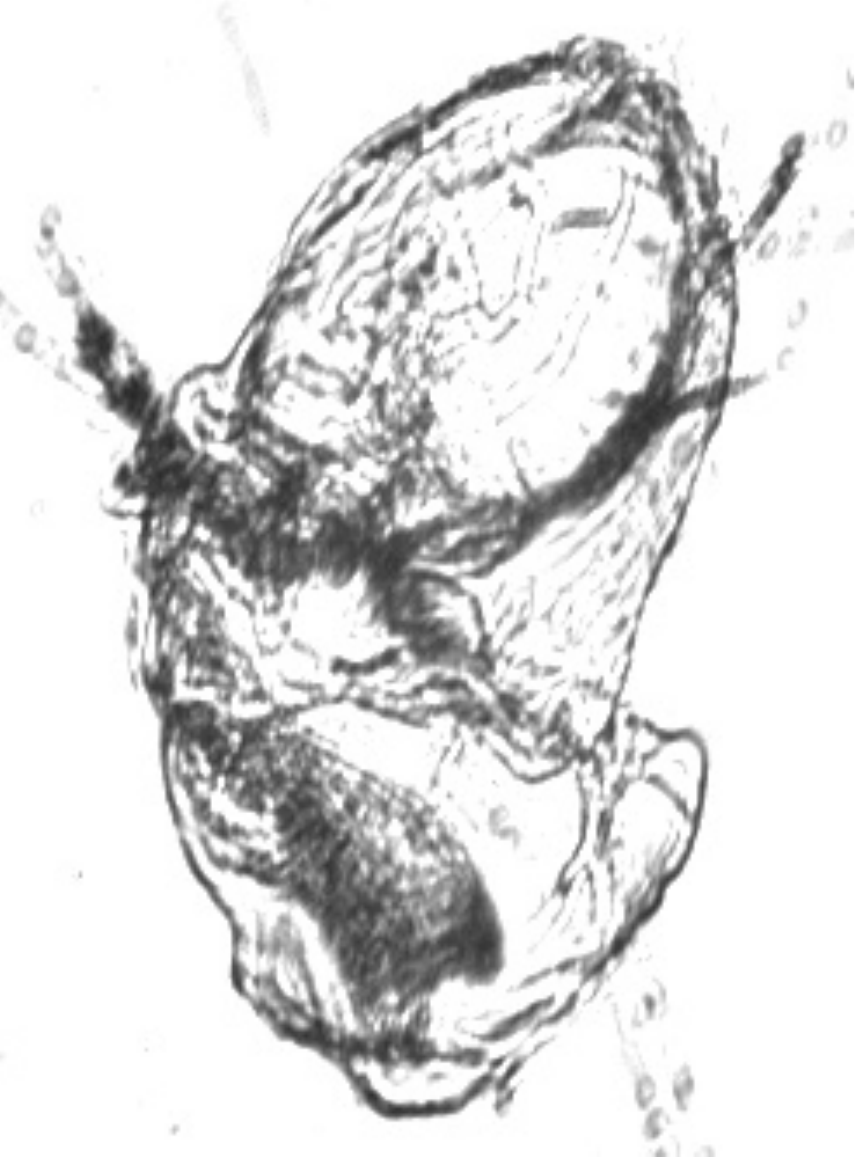}
                \caption{Plankton}
                \label{fig:data-examples-plankton}
        \end{subfigure}%
        \quad\quad 
        \begin{subfigure}[b]{0.16\textwidth}
		\includegraphics[width=\textwidth]{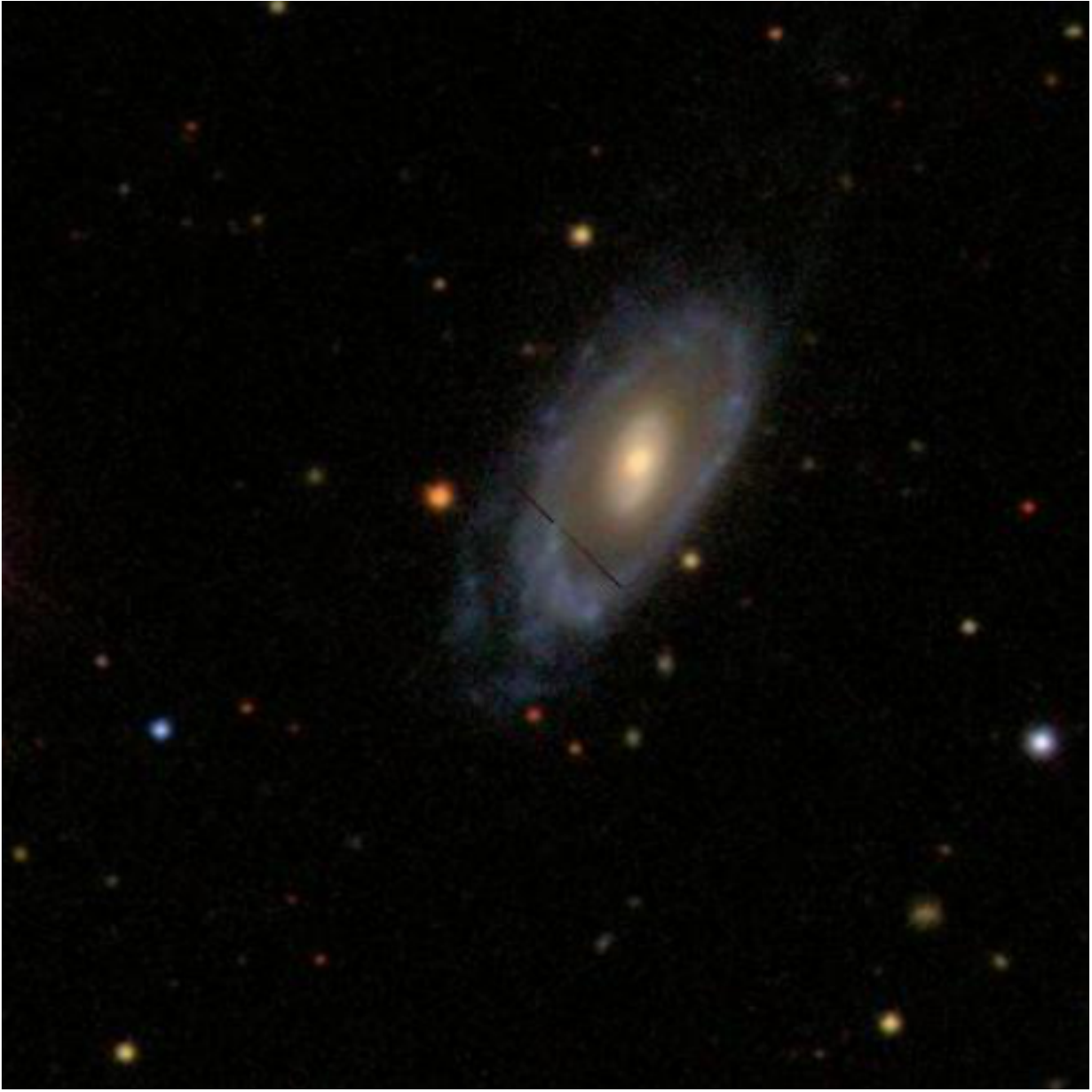}
                \caption{Galaxies}
                \label{fig:data-examples-galaxies}
        \end{subfigure}%
        \caption{Example images for the Plankton and Galaxies datasets, which are rotation invariant.}\label{fig:data-examples}
\end{figure}

By nature of the way the images were acquired, the class of an organism is fully invariant to rotation: ignoring the very minor effects of gravity, the organisms may be arbitrarily oriented when they are photographed. An example image from the training set is shown in Figure \ref{fig:data-examples-plankton}.

\paragraph{Galaxies} The Galaxies dataset consists of 61,578 colour images of size 424$\times$424 for training. We downscaled them by a factor of 4 and cropped them to 64$\times$64. The images display galaxies with various morphological properties and form a subset of the image used in the Galaxy Zoo 2 project \cite{willett2013galaxy}. Each image is classified according to a taxonomy consisting of 11 questions with varying numbers of answers. There are 37 such answers in total. Questions pertain to e.g. the smoothness of the depicted galaxy, whether it has a spiral pattern and how many spiral arms there are. For each image, a vector of 37 probability values corresponding to the answers is provided, estimated from the votes of users of the Galaxy Zoo crowd-sourcing platform\footnote{http://www.galaxyzoo.org/}. We split the dataset into a validation set of 6,157 images and a training set of 55,421 images. Since this dataset was also used for a competition on Kaggle, labels for the test set, containing 79,975 more images, were not provided. We obtained test scores by submitting predictions to Kaggle for this dataset as well.

There is no canonical orientation for galaxies in space, due to the absence of a fixed reference frame. It follows that the morphological properties of a galaxy are independent of the orientation in which we observe it from Earth. This means that the answer probabilities describing these properties should be invariant to rotation of the images. An example image is shown in Figure \ref{fig:data-examples-galaxies}.

\paragraph{Massachusetts buildings} The Massachusetts buildings dataset \cite{MnihThesis} consists of 1500$\times$1500 aerial images of the Boston area, with each image covering an area of 2.25 square kilometers. The dataset features 151 images, with pixelwise annotations of buildings. Following \cite{MnihThesis}, it was split into a training set of 137 images, a validation set of 4 images and a test set of 10 images. During training, we randomly sample smaller 80$\times$80 tiles and predict labels for a 40$\times$40 square in the center.

\begin{figure}
        \centering          
        \begin{subfigure}[b]{0.16\textwidth}
                \includegraphics[width=\textwidth]{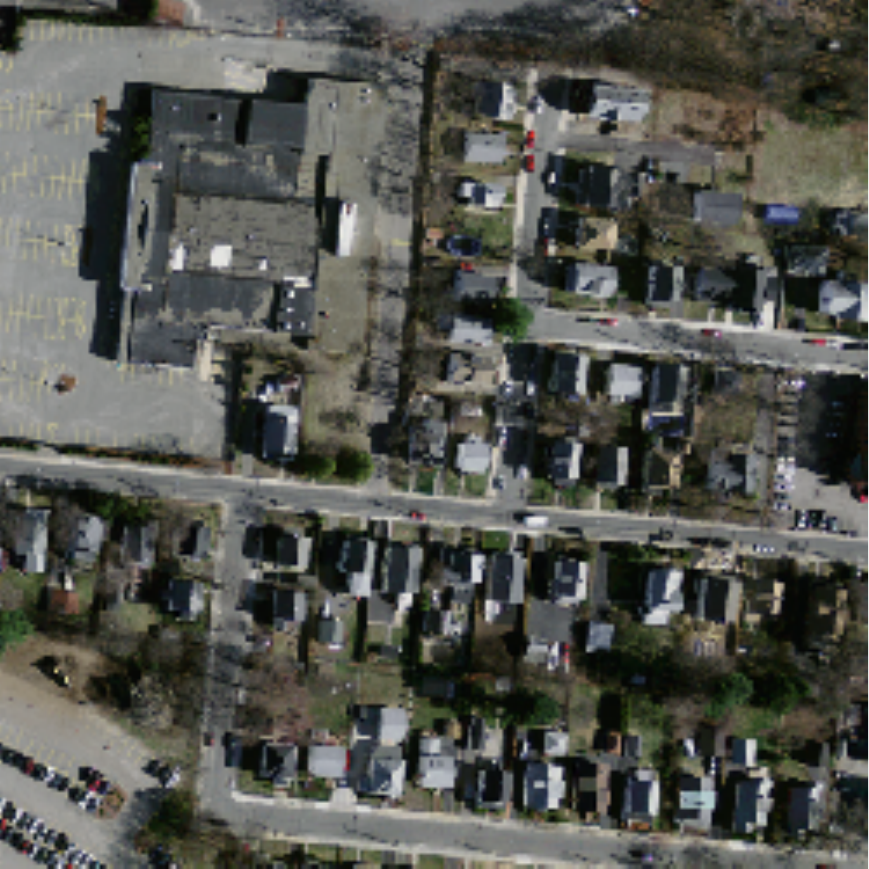}
                \caption{Satellite image}
                \label{fig:data-mass-buildings-x}
        \end{subfigure}%
        \quad\quad 
        \begin{subfigure}[b]{0.16\textwidth}
                \includegraphics[width=\textwidth]{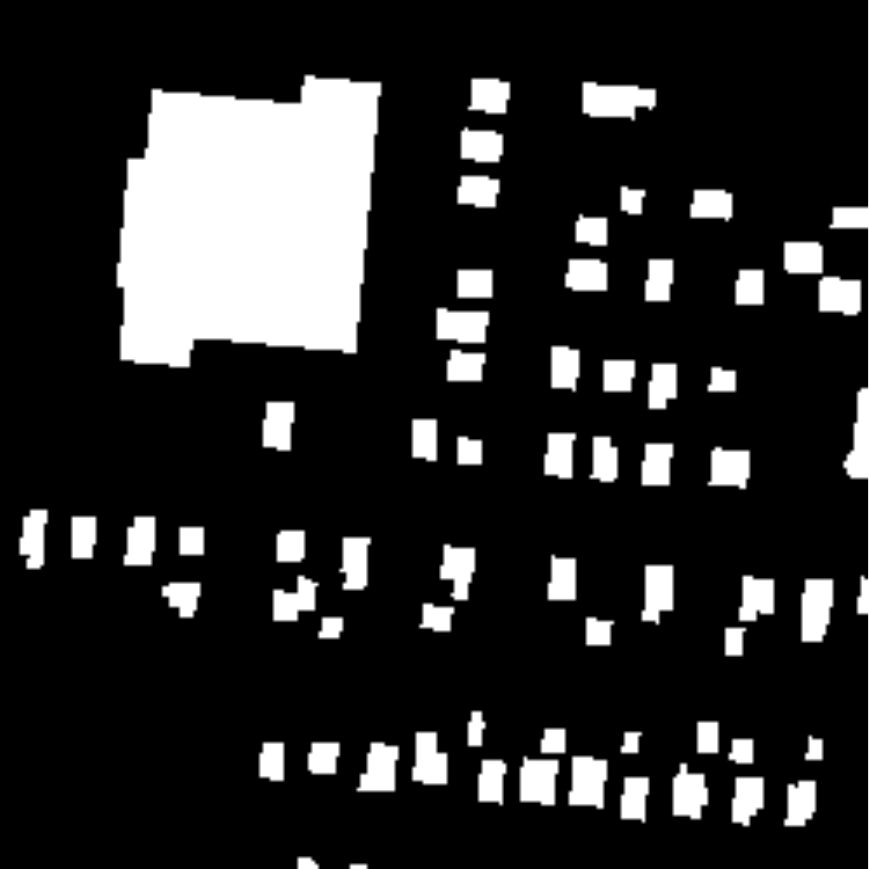}
                \caption{Building labels}
                \label{fig:data-mass-buildings-t}
        \end{subfigure}%
        \caption{Example tile from the Massachusetts buildings dataset, which is same-equivariant to rotation, and corresponding labels.}\label{fig:data-mass-buildings}
\vspace{-10pt}
\end{figure}

Because the annotations for this dataset are pixelwise, rotating an input image should result in an identical rotation of the corresponding output: the task of labeling buildings in satellite images is same-equivariant. An example image and its corresponding label information from the dataset is shown in Figure \ref{fig:data-mass-buildings}.

\subsection{Experimental setup}
\begin{figure}
\begin{center}
\hspace{2em} \includegraphics[width=0.9\columnwidth]{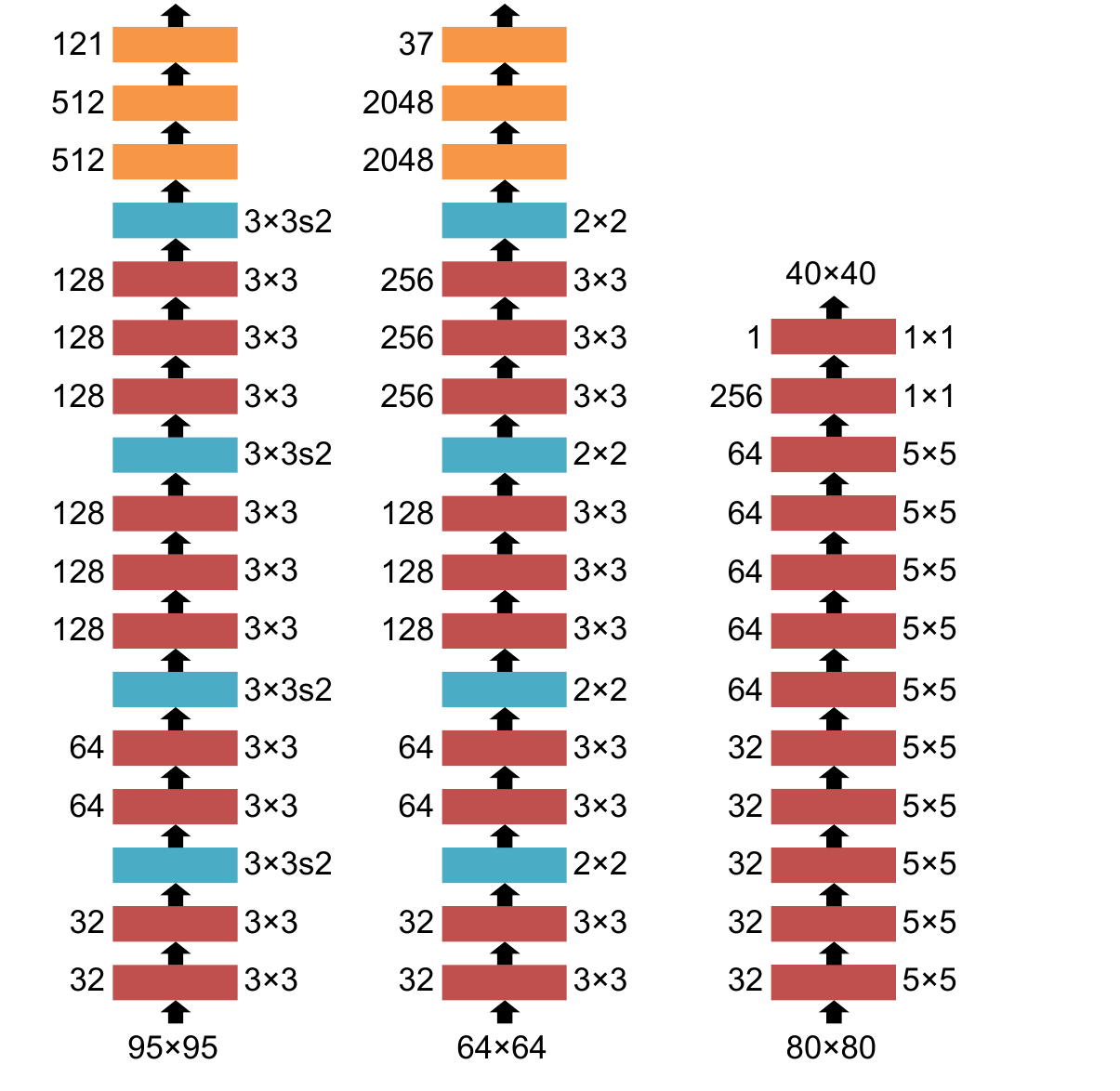}
\caption{Baseline architectures for plankton (left), galaxies (middle) and Massachusetts buildings (right). Conv. layers are shown in red, pooling layers in blue, dense layers in orange. The numbers of units are indicated on the left, filter sizes on the right. ReLUs are used throughout. Dropout with $p=0.5$ is applied before all dense layers.}
\label{fig:baselines}
\vspace{-10pt}
\end{center}
\end{figure}

We use baseline CNNs for each dataset that were designed following today's common practices and achieve competitive performance. For the plankton and galaxies datasets, the architectures are inspired by the VGG architectures \cite{simonyan2014very}, using 3$\times$3 `same' convolutions (which keep the spatial dimensions of the feature maps constant by zero-padding the input) throughout in combination with (overlapping) pooling. These networks would have ranked 12/327 (galaxies) and 57/1050 (plankton) on Kaggle respectively if they had been competition entries, which is quite reasonable when taking into account that top participants used extensive model averaging to get their best results. For the Massachusetts dataset, we use a stack of 5$\times$5 `valid' convolutional layers (which do not pad the input and hence reduce the spatial dimensions) without pooling, followed by 1$\times$1 convolutions, to ensure that enough contextual information is available for each pixel. They are shown in Figure \ref{fig:baselines}.

We use the Adam optimisation method \cite{kingma2014adam} for all experiments, because it allows us to avoid retuning learning rates when cyclic layers are inserted. We use discrete learning rate schedules with tenfold decreases near the end of training, following \citet{Krizhevsky2012}. For the plankton dataset we also use weight decay for additional regularisation. We use data augmentation to reduce overfitting, including random rotation between $0\degree$ and $360\degree$. We made sure to do this
when training the baselines as well, such that they would have the opportunity to learn the desired invariance properties. We focus on cyclic symmetry, because preliminary experiments showed that there is usually no practical benefit to dihedral symmetry. In addition, the eightfold increase in parameter sharing makes it more difficult to compare models on equal footing.

For the plankton dataset, we report the cross-entropy. For the galaxies dataset, we report the prediction root-mean-square error (\emph{RMSE}). For the Massachusetts buildings dataset, we report the area under the ROC curve (\emph{AUC}). We report the mean and standard deviation for these metrics across 10 training runs. We provide them for both the train and test sets, to give an idea of the level of overfitting.

\begin{table}[ht]
\vspace{-10pt}
\caption{Number of model parameters and results on the plankton dataset (cross-entropy, lower is better).}
\label{tab:results-plankton}
\begin{center}
\begin{small}
\begin{tabular}{ccccc}
\hline
\abovespace
\belowspace
\sc{Model}                  & \sc{Size} & \sc{Train} & \sc{Test}               \\ 
\hline
\abovespace
baseline                    & 2.84M          & 0.4064 $\pm$ 0.0044 & 0.7035 $\pm$ 0.0039 \\
\belowspace
baseline \half              & 0.82M          & 0.5770 $\pm$ 0.0059 & 0.7462 $\pm$ 0.0035 \\
\hline 
\abovespace
pool mean                   & 2.84M          & 0.4990 $\pm$ 0.0115 & \textbf{0.6781} $\pm$ 0.0041 \\
pool RMS                    & 2.84M          & 0.5441 $\pm$ 0.0076 & 0.7042 $\pm$ 0.0035 \\
pool max                    & 2.84M          & 0.5385 $\pm$ 0.0127 & 0.7048 $\pm$ 0.0057 \\
mean + ReLU            & 2.84M          & 0.5376 $\pm$ 0.0125 & 0.6982 $\pm$ 0.0062 \\
RMS + ReLU             & 2.84M          & 0.5372 $\pm$ 0.0068 & 0.6935 $\pm$ 0.0027 \\
\belowspace
max + ReLU             & 2.84M          & 0.5686 $\pm$ 0.0075 & 0.7085 $\pm$ 0.0040 \\
\hline
\abovespace 
roll all \quarter           & 0.95M          & 0.5007 $\pm$ 0.0087 & \textbf{0.6961} $\pm$ 0.0052 \\
\belowspace
roll dense \half            & 2.28M          & 0.4868 $\pm$ 0.0106 & \textbf{0.6764} $\pm$ 0.0041 \\
\hline
\end{tabular}
\end{small}
\end{center}
\vskip -0.1in
\end{table}

\begin{table}[ht]
\vspace{-10pt}
\caption{Number of model parameters and results on the galaxies dataset (root-mean-square error, lower is better).}
\vskip -0.12in
\label{tab:results-galaxies}
\begin{center}
\begin{small}
\begin{tabular}{ccccc}
\hline
\abovespace
\belowspace
\sc{Model}                  & \sc{Size} & \sc{Train} & \sc{Test}               \\ 
\hline
\abovespace
baseline                    & 14.57M  & 0.08161 $\pm$ 0.00021 & 0.08343 $\pm$ 0.00019 \\
\belowspace
baseline \half              & 4.75M   & 0.08607 $\pm$ 0.00031 & 0.08710 $\pm$ 0.00025 \\
\hline
\abovespace
\belowspace
pool mean                   & 14.57M  & 0.08023 $\pm$ 0.00017 & \textbf{0.08214} $\pm$ 0.00010 \\
\hline
\abovespace 
roll all \quarter           & 6.84M   & 0.08078 $\pm$ 0.00024 & \textbf{0.08261} $\pm$ 0.00017 \\
\belowspace
roll dense \half            & 14.57M  & 0.08013 $\pm$ 0.00016 & \textbf{0.08200} $\pm$ 0.00009 \\
\hline
\end{tabular}
\end{small}
\end{center}
\vskip -0.1in
\end{table}

\begin{table}[ht]
\vspace{-10pt}
\caption{Number of model parameters and results on the Massachusetts buildings dataset (AUC, higher is better).}
\label{tab:results-mass_buildings}
\begin{center}
\begin{small}
\begin{tabular}{ccccc}
\hline
\abovespace
\belowspace
\sc{Model}                  & \sc{Size} & \sc{Train} & \sc{Test}               \\ 
\hline
\abovespace
baseline                    & 583k    & 0.9774 $\pm$ 0.0005 & 0.9714 $\pm$ 0.0005 \\
\belowspace
baseline \half              & 151k    & 0.9711 $\pm$ 0.0014 & 0.9643 $\pm$ 0.0011 \\
\hline
\abovespace
\belowspace
pool mean                   & 583k    & 0.9806 $\pm$ 0.0012 & \textbf{0.9722} $\pm$ 0.0005 \\
\hline
\abovespace 
roll all \quarter           & 158k    & 0.9773 $\pm$ 0.0011 & \textbf{0.9676} $\pm$ 0.0008 \\
\belowspace
roll all \half              & 598k    & 0.9824 $\pm$ 0.0009 & \textbf{0.9742} $\pm$ 0.0006 \\
\hline
\end{tabular}
\end{small}
\end{center}
\vskip -0.1in
\end{table}

\subsection{Pooling functions}
\label{sec:pooling-functions}
First, we modified the plankton baseline architecture by adding a cyclic slicing layer at the input side, and a cyclic pooling layer just before the output layer. We reduced the batch size used for training by a factor of $4$. We compared three different pooling functions: the \emph{mean}, the maximum (\emph{max}) and the root-mean-square (\emph{RMS}). We also evaluated whether we should apply the \emph{ReLU} nonlinearity to the features before pooling or not. This gives a total of six configurations, which are listed in Table \ref{tab:results-plankton} along with their approximate number of parameters and results.

Mean pooling without a nonlinearity gives the best results in terms of cross-entropy. We will report results using only this pooling function for further experiments, but it should be noted that the choice of function will typically depend on the dataset and the size of the model. Anecdotally, we found that alternative pooling functions sometimes result in better regularization. For the other two datasets, pooling also gives a modest improvement over the baseline.

\subsection{Networks with rolling layers}
Next, we investigated the effect of inserting one or more rolling layers into the networks, in addition to the slicing and pooling layers. We considered two approaches: in one case, we insert rolling layers after all convolutional layers, as well as after the first dense layer (\emph{roll all}), and reduce the number of units in these layers. In the other case, we insert a rolling layer only after the first dense layer (\emph{roll dense}). We can keep the number of feature maps constant for the layers after which rolling operations are inserted by reducing the number of filters by a factor of 4. The layers will then have 4 times fewer parameters. When halving the number of filters instead, the number of feature maps will double relative to the original model. This implies a decrease in parameters for the layer before the rolling operation, but an increase for the next layer.

For all three datasets, we observe a limited effect on performance while the number of parameters is significantly reduced (\emph{roll all \nicefrac{1}{4}}). For each dataset we also report the performance of a version of the baseline network with only half the number of filters in all layers except for the last two (\emph{baseline \nicefrac{1}{2}}), which should have a comparable number of parameters, to demonstrate that the models with rolling layers make more efficient use of the same parameter budget. This is especially interesting because these models take roughly the same amount of computation to train; the additional cost of the roll operations is minimal compared to the cost of the convolutions.

For the plankton and galaxies datasets, the baseline models have several dense layers. The first one of these has the most parameters, because its input consists of a flattened stack of feature maps from the topmost convolutional layer. We can reduce the number of parameters in this layer by halving the number of units and adding a rolling layer (\emph{roll dense \nicefrac{1}{2}}). This doubles the number of parameters of the next dense layer, but for the plankton network the net result is a reduction because that layer had fewer parameters to begin with. Performance is slightly improved compared to the baseline networks.

For the Massachusetts buildings dataset, the baseline model is fully convolutional. We have evaluated a version of the network with roll layers where the number of filters in each layer is reduced only by a factor of 2 (\emph{roll all \nicefrac{1}{2}}), resulting in a network with roughly the same number of parameters as the baseline, but with better performance. Note that for the other datasets, which are more limited in size, such models would heavily overfit.

\section{Conclusion and future work}

We have introduced a framework for building rotation equivariant neural networks, using four new layers which can easily be inserted into existing network architectures. Beyond adapting the minibatch size used for training, no further modifications are required. We demonstrated improved performance of the resulting equivariant networks on datasets which exhibit full rotational symmetry, while reducing the number of parameters. A fast GPU implementation of the rolling operation for Theano (using CUDA kernels) is available at \url{https://github.com/benanne/kaggle-ndsb}.

In future work, we would like to apply our approach to other types of data which exhibit rotational symmetry, particularly in domains where data scarcity is often an issue (e.g. medical imaging), and where additional parameter sharing would be valuable to reduce overfitting. We will also explore the extension of our approach to other groups of transformations, including rotations over angles that are not multiples of $90\degree$, and investigate strategies to manage the additional complexity arising from the required interpolation and realignment of feature maps. Finally, we would like to to extend our work to volumetric data, where reducing the number of parameters is even more important and where a larger number of symmetries can be exploited without requiring costly interpolation. 

\section*{Acknowledgements}
The authors would like to thank Jeroen Burms, Pieter Buteneers, Taco Cohen, Jonas Degrave, Lasse Espeholt, Max Jaderberg, Ira Korshunova, Volodymyr Mnih, Lionel Pigou, Laurent Sifre, Karen Simonyan and A\"aron van den Oord for their insights and input.

\bibliography{paper_arxiv}
\bibliographystyle{icml2016}

\end{document}